\documentclass[conference]{IEEEtran}
\IEEEoverridecommandlockouts % This command is only needed if you want to use the \thanks command

\usepackage{times}
\usepackage[numbers]{natbib}
\usepackage{multicol}
\usepackage[bookmarks=true]{hyperref}
\usepackage{graphicx} % Required for inserting images
\usepackage{newtxtext,newtxmath}
\usepackage{xcolor}
\usepackage{booktabs}
\usepackage[scheme=plain]{ctex}
\usepackage{url}
\usepackage{tabularx}
\usepackage{ragged2e}
\usepackage{makecell}
\usepackage{array}
\usepackage{multirow}
\usepackage{caption}
\usepackage{dblfloatfix}
\usepackage{siunitx}
\usepackage{cuted}

\title{\LARGE \bf
System Design for Maintaining Internal State Consistency in Long-Horizon Robotic Tabletop Games
}

\author{Guangyu Zhao$^{1,2}$, Ceyao Zhang$^{1,2}$, Chengdong Ma$^{1}$, Tao Wu$^{2,3}$, Yiyang Song$^{1,2}$, Haoxuan Ru$^{1,2}$, Yifan Zhong$^{1,2}$, \\
Ruilin Yan$^{2}$, Lingfeng Li$^{1,2}$, Ruochong Li$^{2,4}$, Yu Li$^{1,2}$, Xuyuan Han$^{4}$, Yun Ding$^{4}$, Ruizhang Jiang$^{1,2}$, \\
Xiaochuan Zhang$^{4}$, Yichao Li$^{4}$, Yuanpei Chen$^{1,2,4*}$, Yaodong Yang$^{1,2,4*}$, Yitao Liang$^{1,2,4*}$

\thanks{$^{1}$ Peking University. $^{2}$ PKU-PsiBot Joint Lab. $^{3}$ Nanyang Technological University. $^{4}$ PsiBot.}%
}

\begin{document}
\maketitle
\thispagestyle{empty}
\pagestyle{empty}

\begin{strip}
\begin{minipage}{\linewidth}\centering
\vspace{-30pt}
    \includegraphics[width=\linewidth]{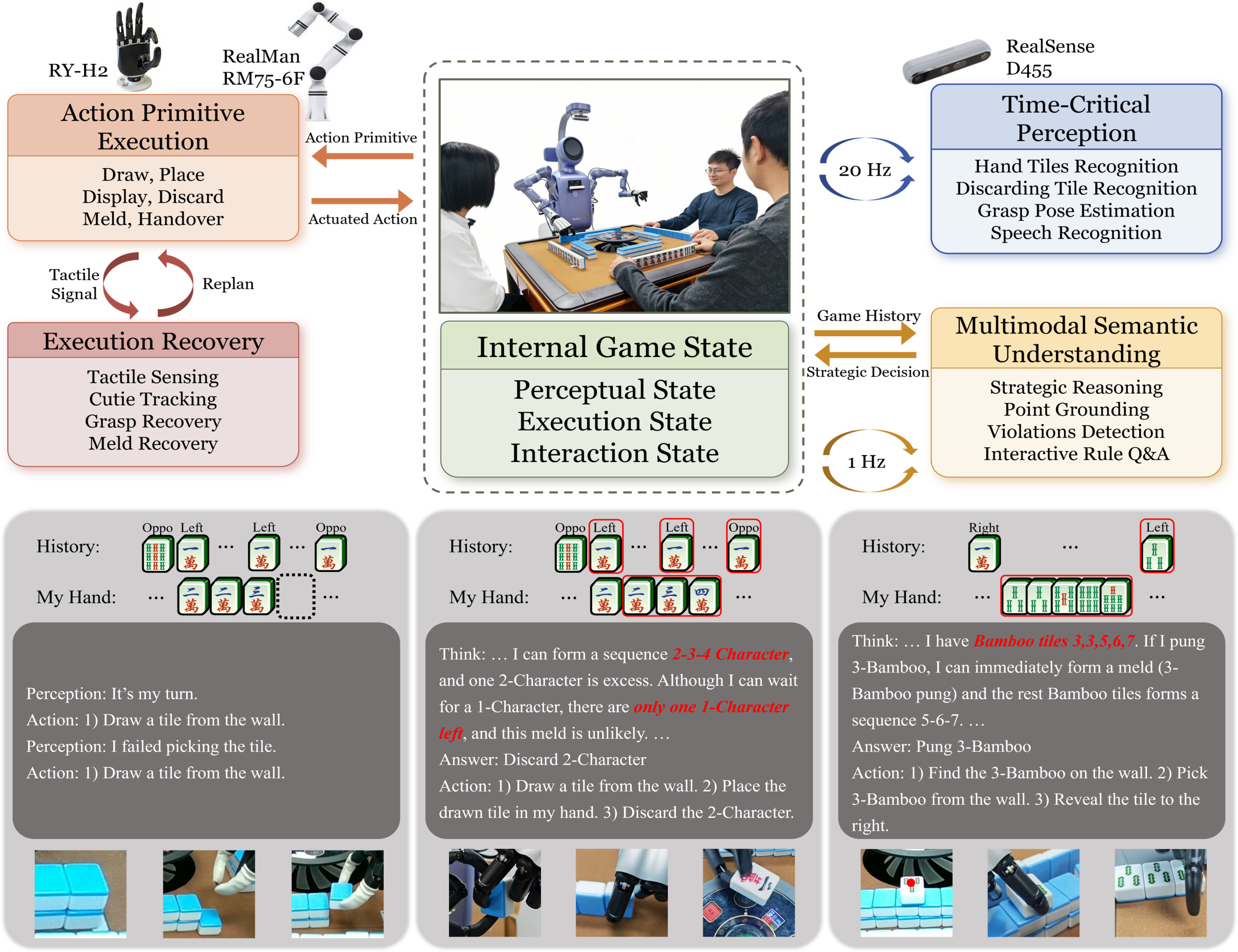}
    \captionof{figure}{\textbf{Overview of the tabletop Mahjong system.} The architecture centers on a maintained internal game state (perceptual, execution, and interaction state). A vision-language model performs strategic reasoning and rule interpretation at low frequency, while time-critical perception modules support real-time detection and pose estimation for manipulation. Action primitives are executed with tactile-based verification and recovery to prevent premature state updates. The bottom examples illustrate how game history informs reasoning and is grounded into verified physical actions within a closed-loop system.}
    \label{fig:overview}
\end{minipage}
\end{strip}

\begin{abstract}
Long-horizon tabletop games pose a distinct systems challenge for robotics: small perceptual or execution errors can invalidate accumulated task state, propagate across decision-making modules, and ultimately derail interaction. This paper studies how to maintain internal state consistency in turn-based, multi-human robotic tabletop games through deliberate system design rather than isolated component improvement. Using Mahjong as a representative long-horizon setting, we present an integrated architecture that explicitly maintains perceptual, execution, and interaction state, partitions high-level semantic reasoning from time-critical perception and control, and incorporates verified action primitives with tactile-triggered recovery to prevent premature state corruption. We further introduce interaction-level monitoring mechanisms to detect turn violations and hidden-information breaches that threaten execution assumptions. Beyond demonstrating complete-game operation, we provide an empirical characterization of failure modes, recovery effectiveness, cross-module error propagation, and hardware–algorithm trade-offs observed during deployment. Our results show that explicit partitioning, monitored state transitions, and recovery mechanisms are critical for sustaining executable consistency over extended play, whereas monolithic or unverified pipelines lead to measurable degradation in end-to-end reliability. The proposed system serves as an empirical platform for studying system-level design principles in long-horizon, turn-based interaction.
\end{abstract}

\IEEEpeerreviewmaketitle

\section{Introduction}

Tabletop games provide a structured yet demanding setting for robotic systems~\cite{openchess, jenga, chinesepokerrobot}. They require repeated manipulation of small objects, tight perception–action loops, strict turn-taking structure, and sustained operation over long horizons, often in the presence of multiple human participants~\citep{gamesmanros}. Mahjong is a particularly challenging instance of this class: gameplay is prolonged, objects are densely arranged, and execution repeatedly alternates between perception, decision-making, and manipulation~\cite{mahjongrobot}. In such settings, small perception or execution errors are not isolated events but can propagate across components, leading to state inconsistency that ultimately derails gameplay~\cite{levine2012monitoring, bohne2023execution}. This makes error recovery and state maintenance a system-level concern rather than a component-level one.

This paper addresses a system-level research question: \emph{what system design choices enable a robotic system to maintain executable state consistency and recoverability during long-horizon, turn-based tabletop interactions with multiple humans?} In physical tabletop games, successful deployment further requires non-trivial game competence: a system that executes actions reliably but behaves indistinguishably from random play is unlikely to be acceptable in practice. Rather than aiming for optimal game-theoretic performance or exhaustive rule enforcement, we focus on sustained, robust operation under realistic physical and interactional conditions, treating executable state consistency as a prerequisite for meaningful, though not necessarily optimal, game performance. Answering this question therefore requires integrating perception, decision-making, manipulation, recovery, and interaction into a single closed-loop system capable of tolerating errors without constant human supervision. Figure~\ref{fig:overview} provides an overview of the complete system.

Early system prototyping revealed deployment-driven constraints that are common in real-world tabletop interactions. Human players often needed clarification of the robot's actions and decisions during ongoing gameplay, and rule-related questions frequently arose mid-game, disrupting turn structure and execution flow. To reduce ambiguity and avoid ad-hoc interventions during long-horizon operation, the system provides two auxiliary interfaces: it records structured decision traces for post-hoc inspection after gameplay, and supports in-game rule queries when questions arise. Rather than treating these deployment-driven observations as human-centered modeling or behavioral research questions, we incorporate them as system-level requirements for deployment.

To address the execution-time validity and robustness demands of physical gameplay, we adopt a hybrid system architecture. A vision–language model is used for high-level reasoning, strategy selection, rule interpretation, and semantic interaction, while specialized perception and control modules handle time-critical tasks such as small-object detection, segmentation, pose estimation, and tactile-based failure detection. For interaction-level perception that depends on game context and turn structure, we found high-level vision--language reasoning to be more reliable in practice than purely visual classifiers, despite lower temporal resolution. This partitioning reflects practical trade-offs observed during system development: while foundation models provide flexibility and semantic understanding, reliable execution with small physical objects requires dedicated components optimized for latency and accuracy. Our evaluation shows that violating this partition—for example by relying on a single model across both semantic reasoning and time-critical perception—leads to measurable degradation in end-to-end system reliability.

Human interaction introduces additional challenges for maintaining internal system state. Rather than attempting comprehensive rule enforcement or adversarial behavior detection, we focus on a minimal set of interaction-level events—specifically, out-of-turn play and unauthorized inspection of other players' hands—that directly violate the robot's execution assumptions and can be handled through detection, alerting, and structured logging. These mechanisms surface potential state inconsistencies to human participants while preserving human authority over rule interpretation and resolution, reflecting the inherently social nature of tabletop games. The system does not interrupt execution flow or attempt to enforce outcomes during gameplay, but instead records sufficient context for post-hoc inspection. Other forms of misconduct, such as direct physical interference with the robot or its objects, are assumed to be mitigated through environmental design and social protocols, as is common in deployed robotic systems.

We evaluate the proposed system across complete games. In addition to system executability, we assess game performance as a deployment-level acceptability signal rather than an optimization objective. The robot completes full games with minimal human intervention (e.g., over 122 games, the robot completed 89.3\% full games without any human intervention) and maintains robust manipulation through tactile-triggered recovery (e.g., 99.8\% grasp success after recovery). We further examine how violations of executable state consistency—such as perception or execution errors that are not successfully recovered—lead to measurable degradation in game outcomes, despite unchanged game strategy. Beyond aggregate success metrics, we report system-level measurements that clarify practical trade-offs in partitioning foundation-model reasoning from time-critical perception and control. These measurements are intended to expose system-level behavior rather than to benchmark individual components in isolation.

Overall, this work contributes: (i) an empirical characterization of failure modes and recovery mechanisms in long-horizon, turn-based tabletop robot games; (ii) evidence that partitioning foundation-model reasoning from time-critical perception and control is critical in practice to sustain state consistency in small-object manipulation tasks; and (iii) a deployed robotic tabletop game system that serves as an empirical vehicle for studying system-level design trade-offs in long-horizon, turn-based human–robot tabletop interactions.

\section{Related Work}

\subsection{Robotic tabletop and board-game systems}
Robotic tabletop games constitute a structured application domain in which perception, manipulation, and interaction must be integrated under strict turn-taking and repeated decision–execution cycles~\cite{chinesepokerrobot, openchess, jenga, mahjongrobot, gamesmanros}. Prior systems demonstrate autonomous or semi-autonomous play in board~\cite{openchess, chessdevelopment} and card~\cite{chinesepokerrobot} games by combining state perception with robotic control, highlighting that rule-governed tabletop interactions place nontrivial demands on physical execution and sensing. Other systems emphasize interaction with human players, balancing game competence with turn structure, pacing, and social acceptability, and often evaluate performance through end-to-end autonomy or user-facing measures rather than isolated component metrics~\cite{zarkowski2019multi,robots-in-games}. Framework-oriented efforts further explore how common perception, decision-making, and actuation components can be reused across multiple tabletop games.

While these systems establish the feasibility of physical game-playing robots, they typically emphasize short-horizon operation or assume reliable state observability during gameplay. In contrast, our work uses Mahjong as a prolonged, turn-based tabletop setting in which any perceptual or execution error can immediately induce internal state inconsistency. Explicit state maintenance and recoverability are central system requirements for sustaining complete-game operation across multiple turns and participants.

\subsection{Partitioning perception, reasoning, and control in integrated robotic systems}

Integrated robotic systems increasingly rely on explicit partitioning between perception, reasoning, and control to balance semantic flexibility with reactive execution, especially in foundation-model-enabled deployments~\citep{fmreview,llmrobotics}. 
Many recent systems ground high-level language or vision--language reasoning in modular perception and control components, translating semantic decisions into executable actions while preserving responsiveness to sensory feedback~\citep{plan-seq-learn,mon2025embodied,fmreview}. 
This architectural separation also provides explicit interfaces for execution monitoring, diagnosis, and localization of failures when system assumptions are violated~\citep{willibald2025hierarchical}. 

Recent approaches study how execution deviations can be detected and corrected at the granularity of skills or task segments, enabling structured recovery without corrupting downstream decision logic~\citep{willibald2025hierarchical}. 
However, many evaluations still emphasize local recovery effectiveness, while the system-level consequences of unrecovered or partially recovered errors across perception, decision-making, and interaction loops remain underexplored in real deployments~\citep{fmreview,mon2025embodied}. 
In turn-based interaction settings, such cross-module propagation can invalidate downstream decisions even when individual actions appear locally successful, motivating explicit system-level state-consistency considerations~\citep{willibald2025hierarchical,mon2025embodied}. 

Recent systems increasingly leverage large language models and vision--language models to expand semantic reasoning and generalization in embodied tasks~\citep{geminirobotics,helix, scaling_helix}.
Some approaches preserve modularity by using foundation models for high-level planning while delegating time-critical perception and control to specialized components~\citep{dexgraspvla, memer}, whereas others pursue more unified mappings from multimodal observations directly to actions~\citep{openvla, pi0, pi05}. 
In practice, this separation becomes particularly important for time-critical loops and precise grounding: grounding-oriented intermediate representations (e.g., masks) can improve low-level manipulation fidelity, while end-to-end VLA models trade modular diagnosability for simplicity~\citep{gondola,roboground, openvla}. 
Our work complements these efforts by examining how deliberate partitioning supports executable state consistency and recovery in long-horizon, turn-based tabletop interaction. Specifically, we study how architectural separation, guarded state transitions, and monitored execution jointly constrain error propagation and preserve internal state validity over extended horizons.

\subsection{Game-playing agents and rule reasoning}

Game-playing agents for board and card games have achieved strong performance by optimizing decision-making under formal rules and well-defined state representations~\citep{alphago,texaspoker,suphx}. In these settings, the game state is typically assumed to be accurately observable or digitally represented, allowing learning and search methods to focus on strategic reasoning and optimal play~\citep{chessgpt, unoarena}. Our focus differs in that game competence is treated as a deployment-level constraint within a physical robotic system operating under imperfect perception, manipulation uncertainty, and multi-human interaction. Rather than optimizing for state-of-the-art game performance or exhaustively enforcing all rule variants, we study how a robotic system can sustain long-horizon execution by maintaining executable state consistency, identifying interaction-level events that threaten system assumptions, and supporting recovery without continuous human intervention.

\section{Problem Formulation: Internal State Consistency in Tabletop Games}

\subsection{Tabletop Games as Long-Horizon Executable Systems}

Tabletop games constitute a distinct class of embodied robotic tasks that differ fundamentally from conventional manipulation or short-horizon interaction scenarios. While individual actions—such as grasping, placing, or pushing objects—may resemble standard tabletop manipulation, gameplay unfolds as a structured execution process governed by turn-taking, persistent state, and repeated interaction with multiple human participants. Each action is interpreted not only by its physical outcome, but also by its implications for the evolving game state, which must remain consistent across perception, decision-making, and interaction over extended horizons.

Three characteristics distinguish tabletop games as long-horizon executable systems. First, turn-taking imposes a strict temporal and causal structure: the validity of an action depends on whose turn it is, what actions have occurred previously, and what state transitions are expected next. Second, gameplay relies on persistent task-level state that must be maintained across turns. Object identities, ownership, and counts (e.g., tiles held by each player) are not transient observations but must be maintained accurately across many turns. Third, tabletop games are inherently multi-human. Human participants act autonomously, may behave unexpectedly, and often rely on implicit social conventions rather than explicit signaling. As a result, interaction-relevant events may occur without explicit machine-readable cues, violating execution assumptions and complicating state maintenance.

In such settings, errors do not remain localized to the component in which they originate. A single perceptual, execution, or interaction error can invalidate shared assumptions about the game state and propagate across modules. For example, misidentifying a tile, missing a state transition due to sensing latency, or failing to execute a grasp can immediately corrupt the robot's internal representation of the game, rendering subsequent decisions invalid even if later perception and control succeed. These failures are often irreversible within the execution logic of the game: an incorrect hand count or turn assumption cannot be corrected without explicit detection and recovery. As a result, sustaining long-horizon operation requires maintaining the consistency of internal state, rather than merely maximizing local action success.

We use Mahjong as a representative instance of this problem class. Mahjong involves dense arrangements of small objects, prolonged turn-based interaction, and repeated transitions between perception, decision-making, and manipulation. Although our system adopts several pragmatic constraints—including simplified scoring rules, and task-specific interaction conventions—these choices preserve the execution structure relevant to long-horizon state maintenance. The focus of this work is therefore not on optimal gameplay or exhaustive rule coverage, but on understanding how a robotic system can maintain executable state consistency under realistic physical and interactional conditions.

\subsection{Definition of Internal State Consistency}

We define \emph{internal state consistency} as the alignment between the robot's maintained internal representation of the game and the actual physical and interaction state of the tabletop environment as gameplay unfolds. The internal state comprises multiple interdependent components, including (i) a perceptual state encoding hand tiles, discarding tile and game history, (ii) an execution state tracking completed and pending actions, and (iii) an interaction state encoding turn structure and human behavior. Consistency requires that these representations remain mutually coherent and grounded in the physical world.

Inconsistencies arise when one or more components diverge from reality or from each other. We distinguish three common classes of inconsistency. \emph{Perceptual inconsistencies} occur when object identities or counts are misestimated or missed due to sensing errors or latency, leading to incorrect updates of the game state. \emph{Execution inconsistencies} arise when physical actions fail or produce unintended outcomes—such as dropped or misplaced objects—causing discrepancies between the assumed and actual state of the game. \emph{Interaction inconsistencies} occur when human actions violate the robot's execution assumptions, for example through out-of-turn play or unauthorized inspection of hidden information, disrupting turn structure and state expectations.

\begin{figure}[htbp]
    \centering
    \includegraphics[width=\columnwidth]{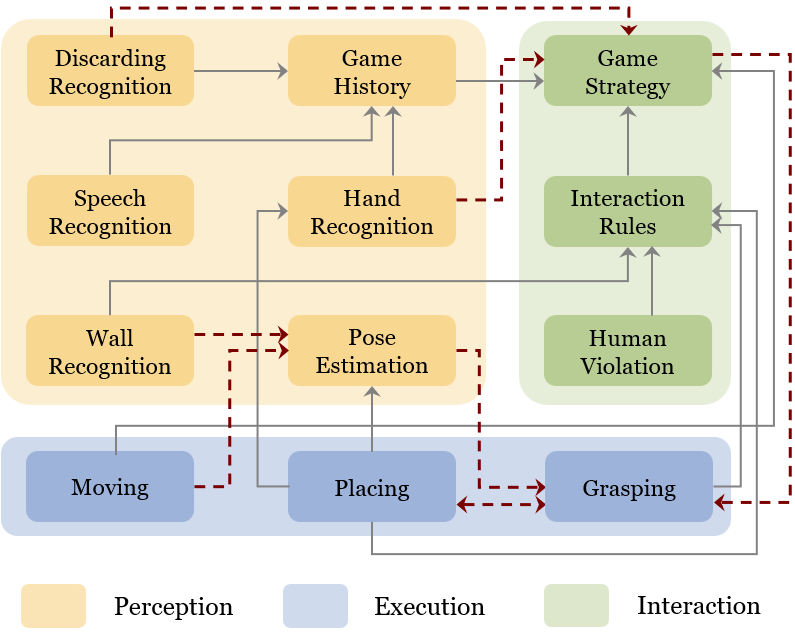}
    \caption{\textbf{Cross-module error propagation in long-horizon tabletop gameplay.} Perceptual, execution, or interaction errors can propagate through the maintained game history and internal state, corrupting downstream reasoning and actions if not detected and recovered. The red dashed lines represent propagation paths that could immediately create visible insecurity.}
    \label{fig:failure_propagation}
\end{figure}

Crucially, these inconsistencies propagate across modules rather than remaining isolated. A perceptual error may corrupt memory, which in turn invalidates downstream strategy reasoning; an execution failure may permanently alter hand counts, eliminating feasible game outcomes; an interaction violation may invalidate turn assumptions, leading to illegal actions. Figure~\ref{fig:failure_propagation} illustrates typical propagation pathways, showing how errors originating in perception, execution, or interaction can cascade through internal state updates and affect subsequent decisions and actions.

This work treats internal state consistency as a system-level property that must be actively maintained and recovered, rather than an implicit assumption. Our system design, monitoring mechanisms, and evaluation are organized around identifying, tolerating, and recovering from state inconsistencies to enable sustained long-horizon gameplay. While certain setup steps (e.g., initial tile arrangement) and task simplifications are handled externally, they do not affect the consistency dynamics during gameplay, which is the focus of our analysis.

\section{System Design for Maintaining State Consistency}

Maintaining state consistency in long-horizon tabletop gameplay requires explicit coordination between perception, reasoning, manipulation, and recovery. In Mahjong, every physical action alters object counts, turn structure, and future decision space. An undetected failure can corrupt the maintained internal state and invalidate subsequent reasoning. Therefore, the key points of the following specific system design all lie in reducing errors, reducing delays, and error recovery.

\textbf{Architectural partitioning.}
The final system deliberately separates high-level semantic reasoning from time-critical recognition and control. A vision–language model~\citep{qwen25vl} fine-tuned by SFT, single-step RL and self-play, performs strategic decision-making, rule interpretation, turn validation, and context-dependent visual judgment. These operations occur at relatively low frequency and operate over the maintained game history. In contrast, latency-sensitive tasks—including tile detection, segmentation, and grasp pose estimation—are handled by specialized components: YOLO~\citep{yolo} for real-time tile detection, SAM~\citep{sam} and FoundationPose~\citep{foundationpose} for pose estimation. Tactile sensing is used for grasp verification. This partition reflects deployment-driven constraints: while foundation models provide semantic flexibility and robustness to contextual variation, small-object detection and repeated manipulation demand deterministic latency and spatial precision. Assigning these responsibilities to separate modules reduces cross-component interference and localizes failure modes within interpretable boundaries.

\textbf{State-conditioned action primitives.}
Gameplay is decomposed into atomic action primitives (draw, place, discard, and meld). Each primitive is executed as a guarded state transition consisting of: (i) precondition validation against the maintained interaction and execution state, (ii) perceptual grounding of the target tile, (iii) pose estimation and motion execution, (iv) post-condition verification via tactile and force feedback, and only then (v) internal state update. Crucially, if execution fails, recovery procedures are triggered without committing state changes. This design prevents premature updates that would otherwise propagate inconsistencies across subsequent turns.

\textbf{Failure detection and recovery.}  
As shown in Figure~\ref{fig:failure_propagation}, physical manipulation errors are the most unsafe source of inconsistency. We found that integrating tactile sensing to detect incomplete grasps during hand closure can reduce the error rate. Upon failure detection, the target tile is re-localized using segmentation tracking~\citep{cutie}, and a new grasp pose is estimated. Throughout this process, internal state variables—such as tile counts and turn progression—remain unchanged. By decoupling execution attempts from state commitment, excecution failures are converted from irreversible state corruption into recoverable events.

\textbf{Interaction-level monitoring.}  
Multi-human gameplay introduces events that violate execution assumptions. Rather than enforcing comprehensive rule compliance, the system monitors two interaction-level conditions that directly threaten internal consistency: out-of-turn play and unauthorized inspection of hidden tiles. These events are evaluated using vision–language reasoning conditioned on the maintained turn state. When detected, they are logged and surfaced to participants without forcibly interrupting gameplay. This preserves human authority in rule interpretation while maintaining transparency regarding potential state inconsistencies. The monitoring mechanism thus functions as a consistency safeguard rather than a rule-enforcement authority.

\section{Results}

\begin{figure}[htbp]
    \centering
    \includegraphics[width=\columnwidth]{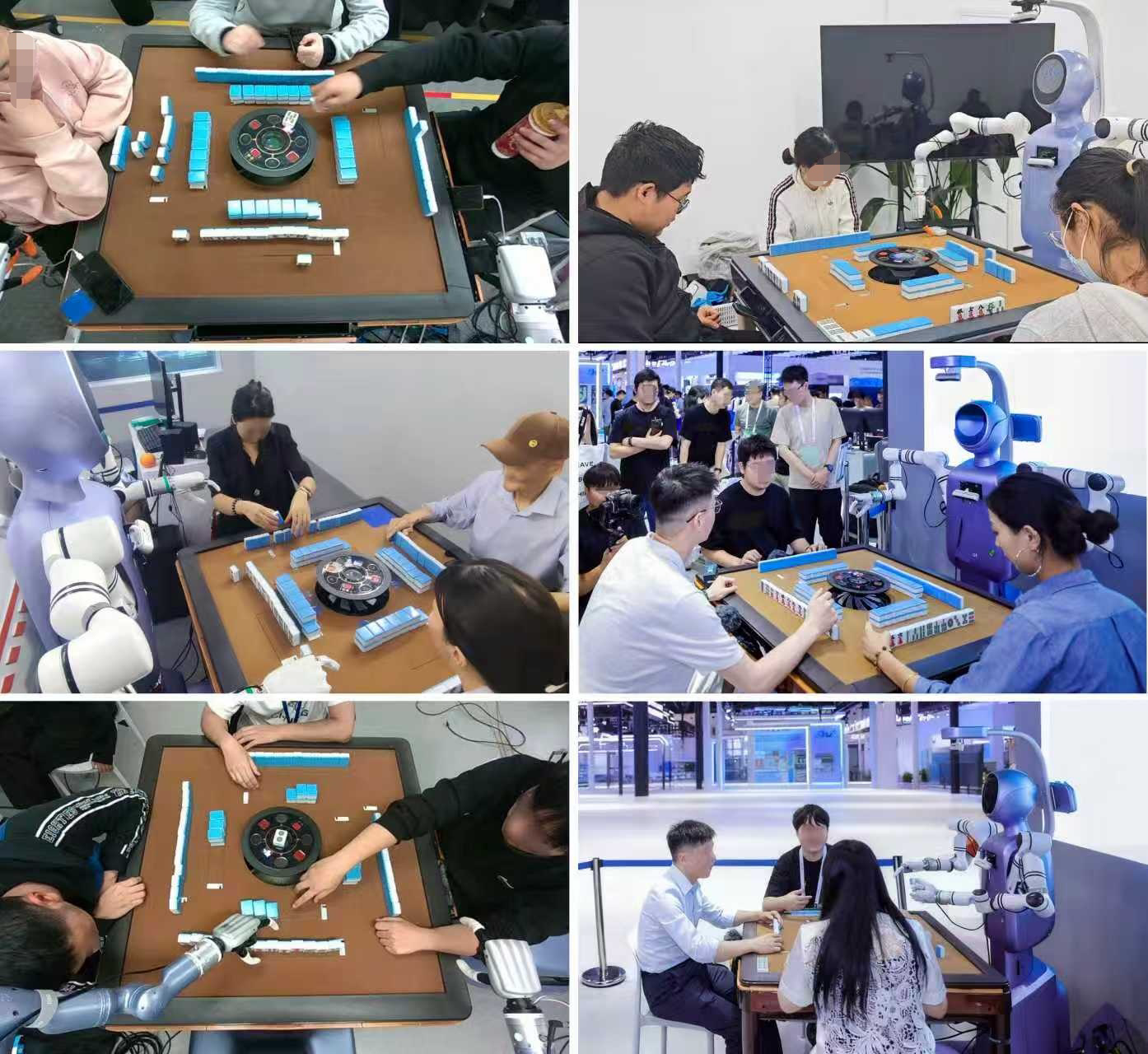}
    \caption{
        \textbf{Representative deployment environments and gameplay settings.} Including
        1) Top-down views of the Mahjong table layout and per-player tile walls.
        2) Close-up of manipulation during discarding and claiming.
        3) Multi-human turn-based gameplay in laboratory conditions.
        4) Public exhibition deployment with multiple participants.
    }
    \label{fig:playing}
\end{figure}

We evaluate the system across complete Mahjong games involving real human participants and autonomous robot operation under sustained, multi-human gameplay conditions. Experiments were conducted in both laboratory and public deployment environments, with multiple participants interacting around a shared tabletop setup, as illustrated in Figure~\ref{fig:playing}. These settings reflect the long-horizon, turn-based interaction context studied in this work and include dense small-object manipulation, imperfect information, and human-driven turn dynamics.

Our evaluation focuses on system-level executability, competitive game performance, perception reliability, and architectural trade-offs. Rather than reporting isolated component benchmarks alone, we present end-to-end measurements that expose how perception, reasoning, execution, and recovery interact over extended horizons, and how failures propagate or are contained within the closed-loop system.

\subsection{Long-Horizon Executability}

Across 122 complete games, the robot successfully sustained gameplay for extended turn sequences involving repeated perception–action loops. Of these, 109 games were completed without any human intervention, corresponding to 89.3\% fully autonomous completion. Of the 13 games that did not complete without intervention, only 3 exhibited unrecoverable failures. The other 10 required only minor human intervention (e.g., manual tile repositioning) and were able to continue to completion thereafter. During the 122 full games, the system executed 2596 action primitives and attempted more than 1000 grasps. The raw single-attempt grasp success rate was 99.2\%. After enabling tactile-triggered recovery, overall grasp success increased to 99.8\%.

\subsection{Game Performance}

Against human players, the robot achieved 34 wins over the 122 games, corresponding to a win rate of 27.9\%, the highest among the four positions on the mahjong table. We adopt paired matches with role reversal to reduce positional bias. For each initial tile configuration, two games are played with swapped seating positions. A paired match is counted as a win only if one agent wins both games; otherwise, it is recorded as a draw. Under this protocol, across 40 paired matches against GPT-5.2, our system achieved 23 wins, GPT-5.2 achieved 3 wins, and 14 paired matches resulted in draws.

Although optimal play is not the primary objective of this work, sustained competitive performance serves as a deployment-level acceptability signal. Notably, win rates degrade measurably when execution inconsistencies remain unrecovered, even though strategic reasoning remains unchanged.

\begin{table*}[t]
\centering
\small
\caption{
\textbf{Human evaluation of decision traces on a Likert scale.}
Fluency measures grammatical correctness and naturalness; 
Independence assesses whether the reasoning appears like human thought rather than tool-assisted; 
Rigor evaluates logical soundness and evidential support (including appropriate use of quantification and certainty); 
Interpretability measures whether the final decision is stepwise explainable and rule-consistent; 
Correctness assesses the strategic quality of the chosen action; 
Overall reflects the perceived human-likeness of the reasoning style. 
We randomly sampled 300 game states from 40 matches and distributed all model outputs evenly to six blinded annotators; five samples were incomplete. 
Values are reported as mean $\pm$ standard deviation, and $n$ denotes the number of evaluated states per model and decision context.
}
\label{tab:trace_eval}

\begin{tabular}{llccccccc}
\toprule
\multirow{2}{*}{Decision Context}
& \multirow{2}{*}{Model}
& \multirow{2}{*}{$n$}
& \multicolumn{6}{c}{Evaluation Metrics (mean $\pm$ std)} \\
\cmidrule(lr){4-9}
& & 
& Fluency 
& Independence 
& Rigor 
& Interpretability 
& Correctness 
& Overall \\
\midrule

\multirow{2}{*}{Discard (Missing Suit)}
& GPT-5.2 & 74 
& 4.23±0.91 & 4.05±1.16 & 4.05±1.08 & 3.82±1.28 & 4.72±0.67 & 3.91±1.09 \\
& Ours & 85 
& \textbf{4.38±1.05} & \textbf{4.27±1.13} & \textbf{4.27±1.15} & \textbf{4.0±1.23} & \textbf{4.85±0.45} & \textbf{4.31±1.03} \\

\midrule
\multirow{2}{*}{Discard (no Missing Suit)}
& GPT-5.2 & 55
& 3.76±1.13 & 3.6±1.43 & 3.53±1.44 & 3.4±1.43 & 3.98±1.39 & 3.49±1.26 \\
& Ours & 44
& \textbf{4.23±1.13} & \textbf{4.27±1.12} & \textbf{4.11±1.21} & \textbf{4.09±1.28} & \textbf{4.57±0.86} & \textbf{4.2±0.94} \\

\midrule
\multirow{2}{*}{Claiming Decision}
& GPT-5.2 & 15
& 3.47±1.02 & 3.0±1.41 & 2.8±1.56 & 2.4±1.58 & 3.13±1.5 & 2.53±1.36 \\
& Ours & 22
& \textbf{4.09±1.04} & \textbf{4.5±0.72} & \textbf{4.23±1.24} & \textbf{4.18±1.03} & \textbf{4.86±0.34} & \textbf{3.95±1.02} \\

\bottomrule
\end{tabular}
\end{table*}

To assess interpretability and post-hoc usefulness of the recorded decision traces, we conducted a human evaluation comparing our system's traces with those generated by GPT-5.2 under identical game states. Human evaluators rated each trace along five dimensions—fluency, naturalness, rigor, correctness, and overall quality. Results are summarized in Table~\ref{tab:trace_eval}.

% \paragraph{Impact of State Inconsistency}

% We further examine games in which execution inconsistencies were not successfully recovered. A total of 8 such games were observed, among which the robot won only 1 game (12.5\% win rate). Although the sample size is limited and does not permit strong statistical claims, the observed trend suggests that unrecovered misalignments—such as incorrect tile counts or turn assumptions—substantially degrade downstream game performance.

\paragraph{Impact of State Inconsistency}

We further examine games in which execution inconsistencies were not successfully recovered. A total of 8 such games were observed, among which the robot won only 1 game (12.5\% win rate). Although the sample size is limited and does not permit strong statistical claims, the observed trend suggests that unrecovered misalignments—such as incorrect tile counts or turn assumptions—substantially degrade downstream game performance.

\begin{table}[t]
\centering
\caption{
Distribution of winning outcomes under two gameplay configurations.
In the \textbf{Characters} mode, used during debugging, the robot records the declared missing suit of all players as Characters.
The \textbf{Normal} mode corresponds to the full system described earlier, where each player's declared missing suit is determined through speech interaction and perception.
Columns \emph{Right}, \emph{Opp.}, and \emph{Left} denote the three human players relative to the robot’s seating position.
Multiple players may win from the same discard, so the sum of wins across players can exceed the number of games.
}
\label{tab:missing_win}
\begin{tabular}{lcccccc}
\toprule
\textbf{Mode} & \textbf{Games} & Robot & Right & Opp. & Left & Draw \\
\midrule
Characters & 46  & 1  & 19 & 14 & 8  & 4 \\
Normal             & 122 & 34 & 29 & 27 & 22 & 13 \\
\bottomrule
\end{tabular}
\end{table}

In addition, during system debugging we occasionally disabled detecting each player's declared missing suit. In these runs, the robot recorded the declared missing suit of all players as Characters for all games. Under this simplified configuration, 46 complete games were played, in which the robot won only once. This contrast suggests that maintaining state consistency in speech interaction and declared-missing-suit perception plays a nontrivial role in overall game performance. Table~\ref{tab:missing_win} summarizes the distribution of winning outcomes under both configurations.

\subsection{Architectural Trade-offs}

Before finalizing the system architecture, we conducted validation experiments to determine the allocation of responsibilities between the vision–language model (VLM) and specialized perception modules.

\textbf{Tile perception.}
For full-frame Mahjong tile recognition, YOLO achieved 98.9\% image-level accuracy with an inference latency of 15$\pm$2 ms per frame. In contrast, VLM-based recognition of individual tiles yielded lower accuracy and required at least 380 ms per image. Given the tight perception–action loop during draw and discard operations, the latency gap alone renders VLM-only detection impractical for sustained gameplay.

\textbf{Interaction-level violation detection.}
We further compared VLM-based reasoning with a DINO-v3~\citep{dinov3} feature extractor followed by a supervised classifier for detecting (i) unauthorized inspection and (ii) turn attribution. Results are summarized in Table~\ref{tab:architecture_tradeoff}. VLM-based reasoning demonstrated higher performance, VLM never misclassified the identity of the inspecting player within positive samples, whereas the DINO-based classifier exhibited a 3.0\% cross-identity error rate.

These findings indicate that time-critical small-object perception benefits from lightweight detectors optimized for latency, whereas interaction-level reasoning requiring contextual interpretation benefits from semantic modeling. The resulting architecture therefore reflects empirical partitioning rather than arbitrary modular design.

\begin{table}[t]
\centering
\caption{
Comparison of architectural alternatives for tile perception and interaction-level violation detection. VLM achieves 94.3\% per-tile recognition accuracy rather than per-image. Latency is measured per image. Precision/Recall are reported for positive-class detection.
}
\label{tab:architecture_tradeoff}
\begin{tabular}{lcc}
\toprule
\textbf{Task} & \textbf{VLM} & \textbf{YOLO / DINO} \\
\midrule
Tile recognition accuracy & $<94.3\%$ & 98.9\% (YOLO) \\
Latency (ms) & $\geq$380 & 15$\pm$2 (YOLO) \\
\midrule
Inspection P/R & 0.967 / 0.989 & 0.956 / 0.960 (DINO) \\
Turn attribution P/R & 0.937 / 0.976 & 0.949 / 0.931 (DINO) \\
Cross-identity error & 0\% & 3.0\% (DINO) \\
\bottomrule
\end{tabular}
\end{table}

\subsection{Perception and Interaction Monitoring Performance}

Across 122 real games, YOLO-based tile recognition produced only 5 misdetections, supporting reliable real-time detection of small objects during gameplay. Turn violation detection achieved a precision of 0.872 and a recall of 0.867, while unauthorized inspection detection achieved a precision of 0.724 and a recall of 0.945. However, the negative predictive value (NPV) and specificity exceed 0.999 for both tasks. This is primarily due to the class imbalance observed during deployment, where negative samples substantially outnumber positive ones, resulting in a data distribution that differs from the training and held-out test sets.

\section{Discussion and Limitations}

\subsection{Failure Case Analysis}

Across all evaluated games, three fatal failures were observed that led to irreversible game termination. One case was caused by an error in speech recognition, which incorrectly detected that a player had declared a winning hand, prematurely terminating the game. The remaining two cases originated from visual misidentification by YOLO: in one instance, a tile was misdetected, leading the system to execute an illegal pung; in another, a misclassification resulted in an illegal concealed kong. These failures illustrate that long-horizon gameplay is vulnerable to cross-module error propagation. In certain cases, a fault in one component can cascade into others, influencing downstream action execution and degrading internal state consistency.

As a result, tactile feedback during grasping and meld execution plays a critical role in preventing such fatal outcomes. When grasp verification or placement confirmation fails, the system delays internal state updates and triggers recovery. Without this verification step, erroneous grasps or unintended tile movements would directly corrupt hand composition, leading to irreversible inconsistencies across subsequent turns.

\subsection{Hardware as a Reliability Bottleneck}

While it is often assumed that system reliability in embodied game playing is primarily limited by algorithmic or software performance, our empirical observations suggest a different picture. Across all recorded failures, 9 incidents were attributable to perception or reasoning errors, whereas 11 failures were associated with hardware-related instability, including grasp closure variance and actuation imprecision.

\begin{figure}[htbp]
    \centering
    \includegraphics[width=\columnwidth]{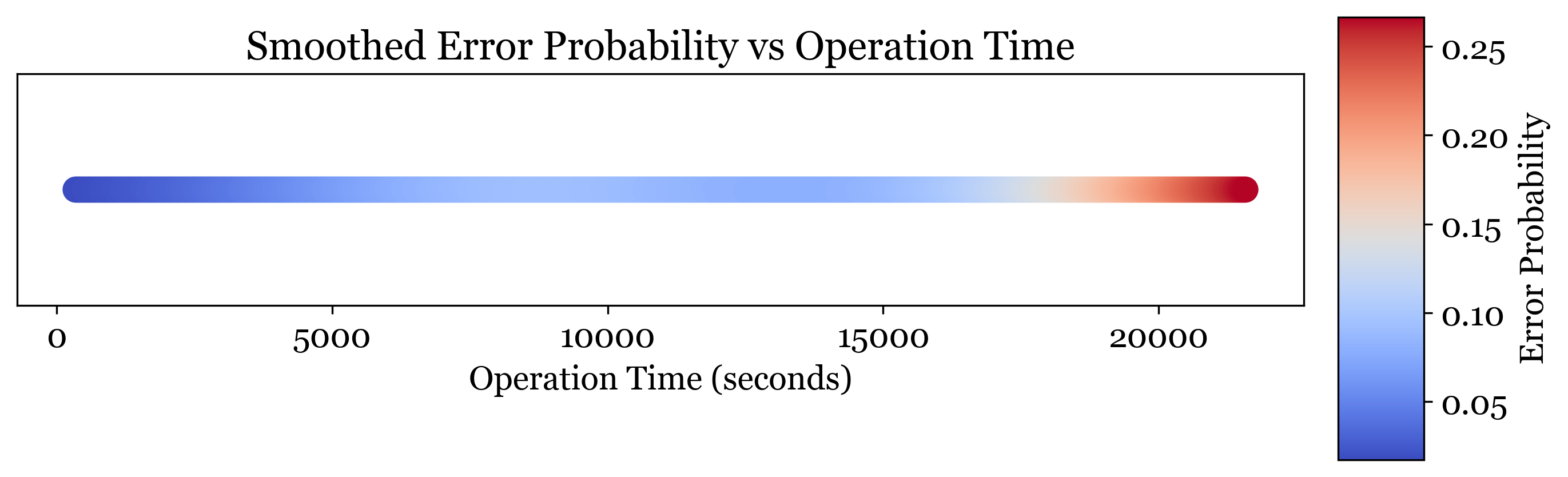}
    \caption{Smoothed hardware error probability as a function of continuous operation time. Error frequency increases noticeably after approximately 20,000 seconds of sustained execution.}
    \label{fig:error_rate}
\end{figure}

To further analyze this effect, Figure~\ref{fig:error_rate} plots the frequency of hardware-induced minor errors as a function of continuous operation time within a single day. We observe a noticeable increase in hardware error rate after approximately 20,000 seconds of sustained operation, indicating performance degradation under prolonged use.

These results suggest that once perception and state estimation reach moderate robustness, mechanical consistency and actuation stability become a dominant factor in end-to-end system reliability, particularly in long-horizon tabletop interaction.

\subsection{Limitations}

The system relies on partially structured environmental assumptions, including calibrated workspace geometry, which limits its adaptability to arbitrary setups. Only a minimal subset of interaction-level violations is monitored, and comprehensive rule enforcement as well as adversarial robustness are beyond the current scope. Furthermore, while the system empirically characterizes failure propagation and recovery, it does not provide formal guarantees of state consistency under all contingencies. The findings should therefore be interpreted as deployment-driven system insights rather than theoretical guarantees.
Although the VLM can generate interpretable reasoning traces that help humans understand the robot’s decisions, its cognitive assistance cannot yet be dynamically adjusted to match user needs. This limitation restricts the robot’s ability to provide progressive assistance for gradual skill development or to reduce challenge when a more relaxed experience is desired. In addition, the extent to which the decision traces generated by large models facilitate post-game reflection or provide meaningful educational value has not yet been systematically evaluated.

\bibliographystyle{plainnat}
\bibliography{references}

@techreport{gamesmanros,
  title       = {GamesmanROS: A Generalized Game-Playing Robotic System},
  author      = {Srikanth, Nikhil},
  institution = {EECS Department, University of California, Berkeley},
  number      = {UCB/EECS-2025-128},
  year        = {2025},
  month       = may,
  note        = {Masters Report. Editors: D. Garcia and K. Goldberg.},
}

@inproceedings{mahjongrobot,
  title     = {A Human-Robot Interactive Mahjong Playing System Based on Visual Recognition Using a Convolutional Neural Network},
  author    = {Wu, Xingcheng and Jing, Siyuan and Tang, Xianwei and Yu, Hao and Shen, Yu and Wang, Ke and Fu, Keren},
  booktitle = {Proceedings of the 2021 5th International Conference on Computer Science and Application Engineering (CSAE '21)},
  year      = {2021},
  pages     = {63:1--63:8},
  publisher = {Association for Computing Machinery},
  doi       = {10.1145/3487075.3487188},
}

@article{openchess,
  title     = {An open-source reproducible chess robot for human-robot interaction research},
  author    = {Zhang, Renchi and de Winter, Joost and Dodou, Dimitra and Seyffert, Harleigh and Eisma, Yke Bauke},
  journal   = {Frontiers in Robotics and AI},
  volume    = {12},
  pages     = {1436674},
  year      = {2025},
  doi       = {10.3389/frobt.2025.1436674},
}

@article{chessdevelopment,
  title={Development of an autonomous chess robot system using computer vision and deep learning},
  author={Phuc, Truong Duc and Son, Bui Cao},
  journal={Results in Engineering},
  volume={25},
  pages={104091},
  year={2025},
  publisher={Elsevier}
}

@article{jenga,
  title     = {Deep Instance Segmentation and Visual Servoing to Play Jenga with a Cost-Effective Robotic System},
  author    = {Marchionna, Luca and Pugliese, Giulio and Martini, Mauro and Angarano, Simone and Salvetti, Francesco and Chiaberge, Marcello},
  journal   = {Sensors},
  volume    = {23},
  number    = {2},
  pages     = {752},
  year      = {2023},
  doi       = {10.3390/s23020752},
  note      = {Also available as arXiv:2211.07977},
}

@article{chinesepokerrobot,
  title={Integration of Robotics, Computer Vision, and Algorithm Design: A Chinese Poker Self-Playing Robot},
  author={Yu, Kuan-Huang},
  journal={arXiv preprint arXiv:2312.09455},
  year={2023}
}

@article{zarkowski2019multi,
  title={Multi-party turn-taking in repeated human--robot interactions: an interdisciplinary evaluation},
  author={{\.Z}arkowski, Mateusz},
  journal={International Journal of Social Robotics},
  volume={11},
  number={5},
  pages={693--707},
  year={2019},
  publisher={Springer}
}

@article{robots-in-games,
  title={Robots in games},
  author={Rato, Diogo and Correia, Filipa and Pereira, Andr{\'e} and Prada, Rui},
  journal={International Journal of Social Robotics},
  volume={15},
  number={1},
  pages={37--57},
  year={2023},
  publisher={Springer}
}

@phdthesis{levine2012monitoring,
  title={Monitoring the execution of temporal plans for robotic systems},
  author={Levine, Steven James},
  year={2012},
  school={Massachusetts Institute of Technology}
}

@article{fmreview,
  title={Real-world robot applications of foundation models: A review},
  author={Kawaharazuka, Kento and Matsushima, Tatsuya and Gambardella, Andrew and Guo, Jiaxian and Paxton, Chris and Zeng, Andy},
  journal={Advanced Robotics},
  volume={38},
  number={18},
  pages={1232--1254},
  year={2024},
  publisher={Taylor \& Francis}
}

@article{llmrobotics,
  title={Large language models for robotics: Opportunities, challenges, and perspectives},
  author={Wang, Jiaqi and Shi, Enze and Hu, Huawen and Ma, Chong and Liu, Yiheng and Wang, Xuhui and Yao, Yincheng and Liu, Xuan and Ge, Bao and Zhang, Shu},
  journal={Journal of Automation and Intelligence},
  volume={4},
  number={1},
  pages={52--64},
  year={2025},
  publisher={Elsevier}
}

@inproceedings{bohne2023execution,
  title={Execution monitoring for long-term autonomous mobile robots in outdoor scenarios},
  author={Bohne, Tim and Kisliuk, Benjamin},
  booktitle={Workshop on Robot Execution Failures and Failure Management Strategies, ICRA},
  year={2023}
}

@inproceedings{plan-seq-learn,
  title={Plan-Seq-Learn: Language Model Guided RL for Solving Long Horizon Robotics Tasks},
  author={Dalal, Murtaza and Chiruvolu, Tarun and Chaplot, Devendra Singh and Salakhutdinov, Ruslan},
  booktitle={ICLR 2024 Workshop on Large Language Model (LLM) Agents},
  year={2024}
}

@article{mon2025embodied,
  title={Embodied large language models enable robots to complete complex tasks in unpredictable environments},
  author={Mon-Williams, Ruaridh and Li, Gen and Long, Ran and Du, Wenqian and Lucas, Christopher G},
  journal={Nature Machine Intelligence},
  volume={7},
  number={4},
  pages={592--601},
  year={2025},
  publisher={Nature Publishing Group UK London}
}

@article{willibald2025hierarchical,
  title={Hierarchical task decomposition for execution monitoring and error recovery: Understanding the rationale behind task demonstrations},
  author={Willibald, Christoph and Lee, Dongheui},
  journal={The International Journal of Robotics Research},
  pages={02783649251352112},
  year={2025},
  publisher={SAGE Publications Sage UK: London, England}
}

@inproceedings{roboground,
  title={Roboground: Robotic manipulation with grounded vision-language priors},
  author={Huang, Haifeng and Chen, Xinyi and Chen, Yilun and Li, Hao and Han, Xiaoshen and Wang, Zehan and Wang, Tai and Pang, Jiangmiao and Zhao, Zhou},
  booktitle={Proceedings of the Computer Vision and Pattern Recognition Conference},
  pages={22540--22550},
  year={2025}
}

@article{gondola,
  title={Gondola: Grounded Vision Language Planning for Generalizable Robotic Manipulation},
  author={Chen, Shizhe and Garcia, Ricardo and Pacaud, Paul and Schmid, Cordelia},
  journal={arXiv preprint arXiv:2506.11261},
  year={2025}
}

@article{memer,
  title={Memer: Scaling up memory for robot control via experience retrieval},
  author={Sridhar, Ajay and Pan, Jennifer and Sharma, Satvik and Finn, Chelsea},
  journal={arXiv preprint arXiv:2510.20328},
  year={2025}
}

@article{dexgraspvla,
  title={Dexgraspvla: A vision-language-action framework towards general dexterous grasping},
  author={Zhong, Yifan and Huang, Xuchuan and Li, Ruochong and Zhang, Ceyao and Chen, Zhang and Guan, Tianrui and Zeng, Fanlian and Lui, Ka Num and Ye, Yuyao and Liang, Yitao and others},
  journal={arXiv preprint arXiv:2502.20900},
  year={2025}
}

@article{pi0,
  title={$\pi_0$: A Vision-Language-Action Flow Model for General Robot Control},
  author={Black, Kevin and Brown, Noah and Driess, Danny and Esmail, Adnan and Equi, Michael and Finn, Chelsea and Fusai, Niccolo and Groom, Lachy and Hausman, Karol and Ichter, Brian and others},
  journal={arXiv preprint arXiv:2410.24164},
  year={2024}
}

@article{pi05,
  title={$\pi_{0.5}$: a Vision-Language-Action Model with Open-World Generalization},
  author={Intelligence, Physical and Black, Kevin and Brown, Noah and Darpinian, James and Dhabalia, Karan and Driess, Danny and Esmail, Adnan and Equi, Michael and Finn, Chelsea and Fusai, Niccolo and others},
  journal={arXiv preprint arXiv:2504.16054},
  year={2025}
}

@article{openvla,
  title={OpenVLA: An Open-Source Vision-Language-Action Model},
  author={Kim, Moo Jin and Pertsch, Karl and Karamcheti, Siddharth and Xiao, Ted and Balakrishna, Ashwin and Nair, Suraj and Rafailov, Rafael and Foster, Ethan and Lam, Grace and Sanketi, Pannag and others},
  journal={arXiv preprint arXiv:2406.09246},
  year={2024}
}

@misc{helix,
  author = {Figure AI},
  title = {Helix: A Vision-Language-Action Model for Generalist Humanoid Control},
  howpublished = {\url{https://www.figure.ai/news/helix}},
  year = {2025}
}

@misc{scaling_helix,
  author = {Figure AI},
  title = {Scaling Helix: a New State of the Art in Humanoid Logistics},
  howpublished = {\url{https://www.figure.ai/news/scaling-helix-logistics}},
  year = {2025}
}

@article{geminirobotics,
  title={Gemini robotics: Bringing ai into the physical world},
  author={Team, Gemini Robotics and Abeyruwan, Saminda and Ainslie, Joshua and Alayrac, Jean-Baptiste and Arenas, Montserrat Gonzalez and Armstrong, Travis and Balakrishna, Ashwin and Baruch, Robert and Bauza, Maria and Blokzijl, Michiel and others},
  journal={arXiv preprint arXiv:2503.20020},
  year={2025}
}

@inproceedings{sam,
  title={Segment anything},
  author={Kirillov, Alexander and Mintun, Eric and Ravi, Nikhila and Mao, Hanzi and Rolland, Chloe and Gustafson, Laura and Xiao, Tete and Whitehead, Spencer and Berg, Alexander C and Lo, Wan-Yen and others},
  booktitle={Proceedings of the IEEE/CVF international conference on computer vision},
  pages={4015--4026},
  year={2023}
}

@inproceedings{cutie,
  title     = {Putting the Object Back into Video Object Segmentation},
  author    = {Cheng, Ho Kei and Oh, Seoung Wug and Price, Brian and Lee, Joon-Young and Schwing, Alexander},
  booktitle = {Proceedings of the IEEE/CVF Conference on Computer Vision and Pattern Recognition (CVPR)},
  year      = {2024},
  note      = {Cutie},
}

@inproceedings{foundationpose,
  title={Foundationpose: Unified 6d pose estimation and tracking of novel objects},
  author={Wen, Bowen and Yang, Wei and Kautz, Jan and Birchfield, Stan},
  booktitle={Proceedings of the IEEE/CVF conference on computer vision and pattern recognition},
  pages={17868--17879},
  year={2024}
}

@inproceedings{yolo,
  title={You only look once: Unified, real-time object detection},
  author={Redmon, Joseph and Divvala, Santosh and Girshick, Ross and Farhadi, Ali},
  booktitle={Proceedings of the IEEE conference on computer vision and pattern recognition},
  pages={779--788},
  year={2016}
}

@article{dinov3,
  title={Dinov3},
  author={Sim{\'e}oni, Oriane and Vo, Huy V and Seitzer, Maximilian and Baldassarre, Federico and Oquab, Maxime and Jose, Cijo and Khalidov, Vasil and Szafraniec, Marc and Yi, Seungeun and Ramamonjisoa, Micha{\"e}l and others},
  journal={arXiv preprint arXiv:2508.10104},
  year={2025}
}

@misc{qwen25vl,
      title={Qwen2.5-VL Technical Report}, 
      author={Shuai Bai and Keqin Chen and Xuejing Liu and Jialin Wang and Wenbin Ge and Sibo Song and Kai Dang and Peng Wang and Shijie Wang and Jun Tang and Humen Zhong and Yuanzhi Zhu and Mingkun Yang and Zhaohai Li and Jianqiang Wan and Pengfei Wang and Wei Ding and Zheren Fu and Yiheng Xu and Jiabo Ye and Xi Zhang and Tianbao Xie and Zesen Cheng and Hang Zhang and Zhibo Yang and Haiyang Xu and Junyang Lin},
      year={2025},
      eprint={2502.13923},
      archivePrefix={arXiv},
      primaryClass={cs.CV},
      url={https://arxiv.org/abs/2502.13923}, 
}

@article{grpo,
  title={Deepseekmath: Pushing the limits of mathematical reasoning in open language models},
  author={Shao, Zhihong and Wang, Peiyi and Zhu, Qihao and Xu, Runxin and Song, Junxiao and Bi, Xiao and Zhang, Haowei and Zhang, Mingchuan and Li, YK and Wu, Yang and others},
  journal={arXiv preprint arXiv:2402.03300},
  year={2024}
}

@article{deepseek-r1,
  title={Deepseek-r1: Incentivizing reasoning capability in llms via reinforcement learning},
  author={Guo, Daya and Yang, Dejian and Zhang, Haowei and Song, Junxiao and Wang, Peiyi and Zhu, Qihao and Xu, Runxin and Zhang, Ruoyu and Ma, Shirong and Bi, Xiao and others},
  journal={arXiv preprint arXiv:2501.12948},
  year={2025}
}

@article{suphx,
  title={Suphx: Mastering mahjong with deep reinforcement learning},
  author={Li, Junjie and Koyamada, Sotetsu and Ye, Qiwei and Liu, Guoqing and Wang, Chao and Yang, Ruihan and Zhao, Li and Qin, Tao and Liu, Tie-Yan and Hon, Hsiao-Wuen},
  journal={arXiv preprint arXiv:2003.13590},
  year={2020}
}

@article{chessgpt,
  title={Chessgpt: Bridging policy learning and language modeling},
  author={Feng, Xidong and Luo, Yicheng and Wang, Ziyan and Tang, Hongrui and Yang, Mengyue and Shao, Kun and Mguni, David and Du, Yali and Wang, Jun},
  journal={Advances in Neural Information Processing Systems},
  volume={36},
  pages={7216--7262},
  year={2023}
}

@article{alphago,
  title={Mastering the game of Go with deep neural networks and tree search},
  author={Silver, David and Huang, Aja and Maddison, Chris J and Guez, Arthur and Sifre, Laurent and Van Den Driessche, George and Schrittwieser, Julian and Antonoglou, Ioannis and Panneershelvam, Veda and Lanctot, Marc and others},
  journal={nature},
  volume={529},
  number={7587},
  pages={484--489},
  year={2016},
  publisher={Nature Publishing Group UK London}
}

@article{texaspoker,
  title={Superhuman AI for multiplayer poker},
  author={Brown, Noam and Sandholm, Tuomas},
  journal={Science},
  volume={365},
  number={6456},
  pages={885--890},
  year={2019},
  publisher={American Association for the Advancement of Science}
}

@inproceedings{unoarena,
  title={UNO arena for evaluating sequential decision-making capability of large language models},
  author={Qin, Zhanyue and Wang, Haochuan and Liu, Deyuan and Song, Ziyang and Fan, Cunhang and Lv, Zhao and Wu, Jinlin and Lei, Zhen and Tu, Zhiying and Chu, Dianhui and others},
  booktitle={Proceedings of the 2024 Conference on Empirical Methods in Natural Language Processing},
  pages={7630--7645},
  year={2024}
}

@article{ppo,
  title={Proximal policy optimization algorithms},
  author={Schulman, John and Wolski, Filip and Dhariwal, Prafulla and Radford, Alec and Klimov, Oleg},
  journal={arXiv preprint arXiv:1707.06347},
  year={2017}
}

@article{dpo,
  title={Direct preference optimization: Your language model is secretly a reward model},
  author={Rafailov, Rafael and Sharma, Archit and Mitchell, Eric and Manning, Christopher D and Ermon, Stefano and Finn, Chelsea},
  journal={Advances in neural information processing systems},
  volume={36},
  pages={53728--53741},
  year={2023}
}

@article{hybridflow,
  title   = {HybridFlow: A Flexible and Efficient RLHF Framework},
  author  = {Guangming Sheng and Chi Zhang and Zilingfeng Ye and Xibin Wu and Wang Zhang and Ru Zhang and Yanghua Peng and Haibin Lin and Chuan Wu},
  year    = {2024},
  journal = {arXiv preprint arXiv: 2409.19256}
}

\appendix

\subsection{Training Pipeline for Strategic Reasoning}
\label{sec:appendix_training}

\begin{figure*}[htbp]
    \centering
    \includegraphics[width=\linewidth]{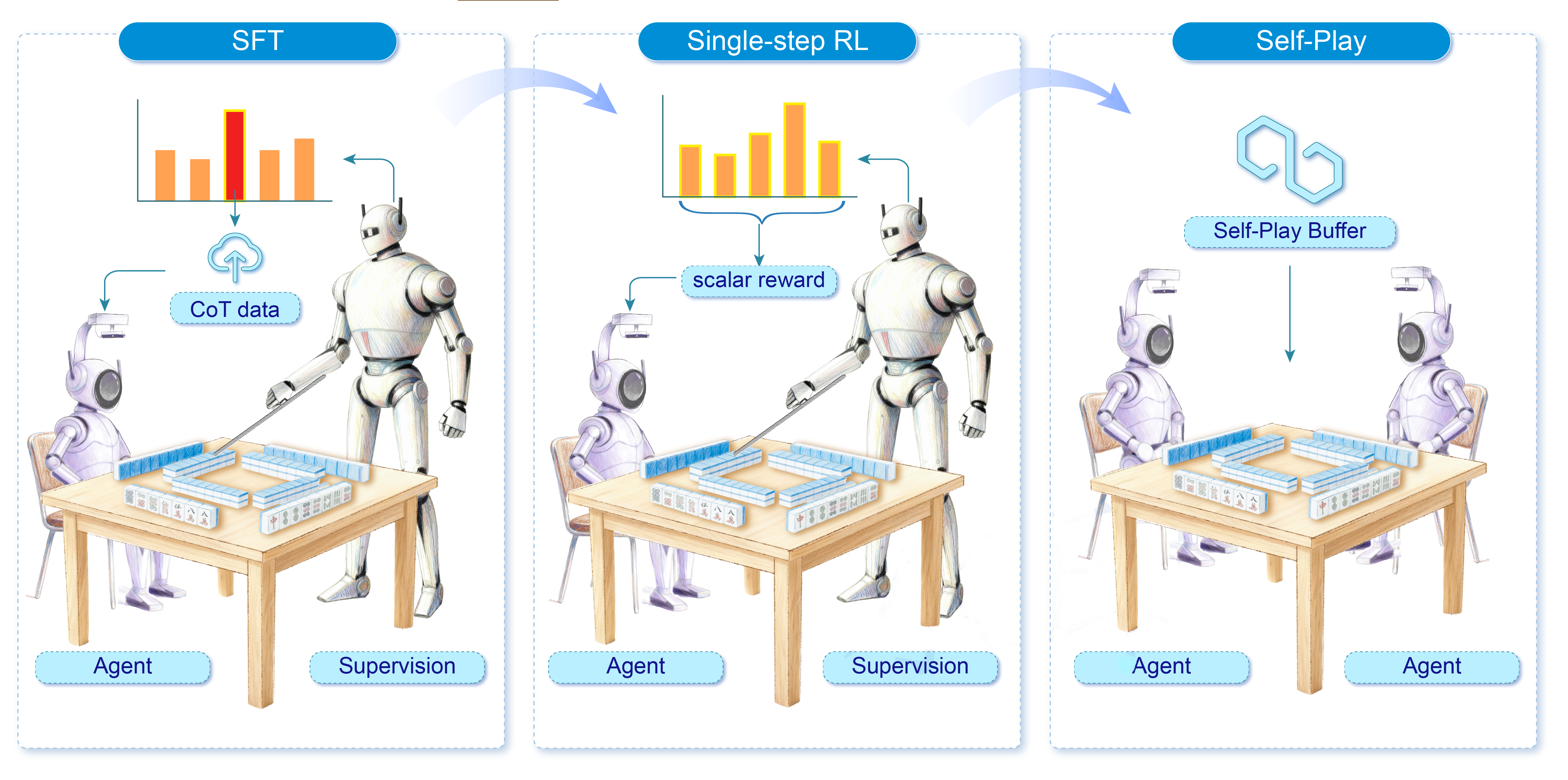}
    \caption{
        \textbf{Training pipeline for strategic reasoning.}
        Stage~1 distills a conventional RL-trained policy into the VLM via supervised fine-tuning with LLM-synthesized reasoning traces.
        Stage~2 applies single-step RL (GRPO) to optimize decision quality beyond imitation.
        Stage~3 uses self-play with DPO to discover strategies that surpass the original teacher policy.
    }
    \label{fig:training}
\end{figure*}

The multimodal semantic understanding module in Figure~\ref{fig:overview} employs Qwen-2.5-VL-7B~\citep{qwen25vl} as the backbone vision--language model (VLM).
For strategic reasoning, the model receives a natural language description of the current game state and history, and produces both an explicit chain-of-thought rationale and a final action decision.
We developed a three-stage training pipeline---supervised fine-tuning (SFT), single-step reinforcement learning (RL), and self-play optimization---that progressively improves game performance while preserving human-readable reasoning traces.
The overall pipeline is illustrated in Figure~\ref{fig:training}.

%--------------------------------------------------------------
\subsubsection*{Stage 1: Supervised Fine-Tuning}
%--------------------------------------------------------------

We first train a conventional MLP-and-CNN-based Mahjong policy $\pi_{\text{AI}}$ using Proximal Policy Optimization (PPO)~\citep{ppo} via self-play.
This teacher policy generates a dataset of state--action pairs $\mathcal{D}_0 = \{(s_i, a_i)\}_{i=1}^{N}$, where each state $s_i$ encodes the player's hand tiles, visible discards, and inferred missing-suit information, and $a_i$ is the action selected by $\pi_{\text{AI}}$.

To equip the VLM with explicit reasoning capabilities, we use DeepSeek-R1~\citep{deepseek-r1} to synthesize chain-of-thought (CoT) explanations for each state--action pair.
Given $(s_i, a_i)$, the LLM generates a natural language reasoning trace $T_i$ that articulates the strategic considerations behind action $a_i$.
This yields an augmented dataset $\widetilde{\mathcal{D}} = \{(s_i, T_i, a_i)\}_{i=1}^{N}$.

We fine-tune the VLM on $\widetilde{\mathcal{D}}$ by minimizing the standard negative log-likelihood:
\begin{equation}
    \mathcal{L}_{\text{SFT}}(\theta) = -\,\mathbb{E}_{(x,\, T,\, a)\,\sim\,\widetilde{\mathcal{D}}} \!\Big[\, \log\, p_{\theta}(T, a \mid x) \,\Big],
\end{equation}
where $\theta$ denotes the VLM parameters.
The resulting model $\pi_{\text{VLM}}^{\text{SFT}}$ learns to produce well-formatted, reasoned outputs.
When evaluated against $\pi_{\text{AI}}$ in four-player games, $\pi_{\text{VLM}}^{\text{SFT}}$ achieves a win rate of 28\%.

%--------------------------------------------------------------
\subsubsection*{Stage 2: Single-Step Reinforcement Learning}
%--------------------------------------------------------------

The SFT stage teaches the model to generate structured, human-readable outputs, but the maximum-likelihood objective treats all tokens equally and does not directly optimize for decision quality.
To improve gameplay performance, we apply GRPO~\citep{deepseek-r1} by VERL~\citep{hybridflow} as a single-step RL procedure, using $\pi_{\text{AI}}$ as a reward signal.

For each state $s$ (described by prompt $x$), the current policy generates $G$ candidate responses $\{(T_i, a_i)\}_{i=1}^{G}$ in parallel.
Each response is scored by a composite reward:
\begin{equation}
    R_i = R_{\text{format}}(T_i, a_i) + \, R_{\text{accuracy}}(s, a_i),
\end{equation}
where $R_{\text{format}}(T_i, a_i) \in \{0, 1\}$ indicates whether the output satisfies predefined formatting constraints (parseable structure, valid action token), $R_{\text{accuracy}}(s, a_i) = \pi_{\text{AI}}(a_i \mid s)$ is the probability assigned by the teacher policy to the chosen action.

The group-level advantage for the $i$-th response is computed as:
\begin{equation}
    A_i = \frac{R_i - \mu_{\mathcal{G}}}{\sigma_{\mathcal{G}}},
\end{equation}
where $\mu_{\mathcal{G}}$ and $\sigma_{\mathcal{G}}$ are the mean and standard deviation of rewards within the group of $G$ responses.

Denoting the importance-sampling ratio $r_i(\theta) = \frac{\pi_\theta(T_i, a_i \mid x)}{\pi_{\text{ref}}(T_i, a_i \mid x)}$ with reference policy $\pi_{\text{ref}} = \pi_{\text{VLM}}^{\text{SFT}}$, the GRPO objective is:
\begin{equation}
\begin{split}
    \mathcal{L}_{\text{GRPO}}(\theta) 
    = -\,\frac{1}{G}\sum_{i=1}^{G}
    \min\!\Big(\,
        r_i(\theta)\, A_i,\; \\
        \mathrm{clip}\big(r_i(\theta),\, 1{-}\varepsilon,\, 1{+}\varepsilon\big)\, A_i
    \,\Big)
    + \beta\, D_{\mathrm{KL}}\!\big[\pi_\theta \| \pi_{\text{ref}}\big]
\end{split}
\end{equation}
where $\varepsilon$ is the clipping threshold and $\beta$ controls the KL penalty strength.

During this stage, the model generates rationales directly from the game state without relying on pre-written traces, and rewards are computed solely based on the final action and output format.
The resulting model $\pi_{\text{VLM}}^{\text{RL}}$ achieves a win rate of 44\% against $\pi_{\text{AI}}$, a substantial improvement over the SFT baseline.

%--------------------------------------------------------------
\subsubsection*{Stage 3: Self-Play Optimization}
%--------------------------------------------------------------

To enable the model to discover strategies beyond those captured by the teacher policy $\pi_{\text{AI}}$, we perform self-play optimization using Direct Preference Optimization (DPO)~\citep{dpo}.

We organize self-play games into groups of $G$ games, where all games within a group share the identical initial wall arrangement and starting hands for all four players.
Divergent trajectories therefore arise solely from different decision sequences.
After completing all $G$ games in a group, we construct a trie over their trajectories: each node corresponds to a state--action pair, and each root-to-leaf path represents one complete game.
Every node records its empirical win rate across all games that pass through it.

We extract preference pairs from nodes that have multiple children, i.e., states where $\pi_{\text{VLM}}^{\text{RL}}$ selected different actions across games.
Children are ranked by their empirical win rates; higher-win-rate actions are treated as preferred and lower-win-rate actions as dispreferred.
For each such contrastive pair, we sample one reasoning trace from each child node, yielding a preferred response $(T^+, a^+)$ and a dispreferred response $(T^-, a^-)$ conditioned on the same state description $x$.

The model is then optimized with the DPO loss:
\begin{equation}
\begin{split}
    \mathcal{L}_{\text{DPO}}(\theta) 
    = -\log\sigma\!\Bigg(
        \beta_{\text{sp}} \bigg[
            \log \frac{\pi_\theta(T^+, a^+ \mid x)}
                      {\pi_{\text{VLM}}^{\text{RL}}(T^+, a^+ \mid x)} \\
            - \log \frac{\pi_\theta(T^-, a^- \mid x)}
                        {\pi_{\text{VLM}}^{\text{RL}}(T^-, a^- \mid x)}
        \bigg]\ 
    \!\Bigg)
\end{split}
\end{equation}
where $\pi_{\text{VLM}}^{\text{RL}}$ serves as the reference policy and $\beta_{\text{sp}}$ controls the KL penalty strength, ensuring the model retains reasoning capabilities acquired in earlier stages.

Through iterative rounds of self-play data collection and DPO updates, the final model achieves a win rate of 48\% against the teacher policy $\pi_{\text{AI}}$.
This demonstrates that the self-play stage enables the VLM to surpass its original teacher, discovering strategies that were not present in the distillation data.

% \input{appendix/resources}
% 代码开源：没有 ROS + 用 ROS 两个版本
% VLM 开源
% prompt 开源
% 全部 YOLO 检测、牌墙头部、放置数据集开源

\end{document}